\documentclass[journal,twoside,web]{ieeecolor}
\usepackage{generic}
\usepackage{cite}
\usepackage{amsmath,amssymb,amsfonts}
\usepackage{algorithmic}
\usepackage{graphicx}
\usepackage{algorithm}

\usepackage{textcomp}
\usepackage{multirow}
\usepackage{booktabs}

\usepackage{balance}
\usepackage{mathrsfs}
\usepackage{indentfirst}

\def\BibTeX{{\rm B\kern-.05em{\sc i\kern-.025em b}\kern-.08em
    T\kern-.1667em\lower.7ex\hbox{E}\kern-.125emX}}
\begin{document}
\title{Towards Safer Robot-Assisted Surgery: \\A Markerless Augmented Reality Framework}
\author{Ziyang Chen, Laura Cruciani, Ke Fan, Matteo Fontana, Elena Lievore, Ottavio De Cobelli, \\ Gennaro Musi, Giancarlo Ferrigno, and Elena De Momi
\thanks{Ziyang Chen, Laura Cruciani, Ke Fan, and Giancarlo Ferrigno are with the Department of Electronics, Information and Bioengineering, Politecnico di Milano, 20133, Italy (e-mail: ziyang.chen@polimi.it; laura.cruciani@mail.polimi.it; ke.fan@polimi.it; giancarlo.ferrigno@polimi.it).}
\thanks{Matteo Fontana, and Elena Lievore are with the Department of Urology, European Institute of Oncology, IRCCS, Milan, 20141, Italy (e-mail: matteo.fontana@ieo.it; elena.lievore@ieo.it).}
\thanks{Ottavio De Cobelli, and Gennaro Musi are with the Department of Urology, European Institute of Oncology, IRCCS, Milan, 20141, Italy, and also with the Department of Oncology and Onco-haematology, Faculty of Medicine and Surgery, University of Milan, Milan, Italy (e-mail: ottavio.decobelli@ieo.it; gennaro.musi@ieo.it).}
\thanks{Elena De Momi is with the Department of Electronics, Information and Bioengineering, Politecnico di Milano, 20133, Italy, and also with European Institute of Oncology, IRCCS, Milan, 20141, Italy (e-mail: elena.demomi@polimi.it).}}

\maketitle

\begin{abstract}

Robot-assisted surgery is rapidly developing in the medical field, and the integration of augmented reality shows the potential of improving the surgeons' operation performance by providing more visual information. In this paper, we proposed a markerless augmented reality framework to enhance safety by avoiding intra-operative bleeding which is a high risk caused by the collision between the surgical instruments and the blood vessel. Advanced stereo reconstruction and segmentation networks are compared to find out the best combination to reconstruct the intra-operative blood vessel in the 3D space for the registration of the pre-operative model, and the minimum distance detection between the instruments and the blood vessel is implemented. A robot-assisted lymphadenectomy is simulated on the da Vinci Research Kit in a dry lab, and ten human subjects performed this operation to explore the usability of the proposed framework. The result shows that the augmented reality framework can help the users to avoid the dangerous collision between the instruments and the blood vessel while not introducing an extra load. It provides a flexible framework that integrates augmented reality into the medical robot platform to enhance safety during the operation.
\end{abstract}


\section{INTRODUCTION}
\IEEEPARstart{R}obot-assisted surgery (RAS) got improved patient outcomes in both intra-operative operation and post-operative recovery compared to traditional open surgery, and it also provides the possibility to integrate artificial intelligence in this platform for autonomous and safe operation. An emerging technology, i.e., Augmented Reality (AR) fusing virtual targets on real scenes, provides more visual information for the users, and it has been introduced in the field of robotic surgery to enhance the safety \cite{hu2021head, chen2023frsr, hussain2019contribution}. Generally, the surgeon needs to capture the pre-operative images of the patient such as Computed Tomography (CT) and Magnetic Resonance Imaging (MRI) for the preliminary surgical planning. These CT/MRI slices can be segmented using software like 3D Slicer to generate the pre-operative 3D model, and then projected on the intra-operative images to implement the AR effect. These augmented intra-operative images can guide the operation of the surgeons by providing more visual information and improve the operation performance. One major challenge is how to localize the intra-operative soft tissues or organs so that the pre-operative model can register with the intra-operative target correctly, to implement the overlapping of the pre-operative model on the corresponding position of intra-operative images. In~\cite{bianchi2021use, schiavina2021real}, the authors implemented the AR in robot-assisted radical prostatectomy by overlapping the pre-operative 3D model on the endoscopic images, they used the software vMIX (StudioCoast Pty Ltd, Australia) to manually align the position of the pre-operative model, which hinders the practice in real-time AR visualization during the operation. Similarly, a manual alignment between the pre-operative model and the intra-operative anatomy was used in some other operations, such as the robotic thyroidectomy~\cite{lee2018preliminary}, the transoral surgery~\cite{chan2020augmented} and the partial nephrectomy~\cite{de2023improving}. In~\cite{wendler2021molecular}, the authors introduced three possible solutions to implement AR registration including landmarks, laparoscopic video and intra-operative ultrasound to recognize the 3D position of surgical scenes, and they summarized that the laparoscopic video-based approach will be the mainstream since it does not require external hardware. Also, the authors in~\cite{qian2019review} introduced the registration process using the robotic instruments by pointing at pre-installed markers or the extra projector-camera system. To implement the automatic AR alignment using the laparoscopic video based on the da Vinci Surgical System (dVSS), the authors in~\cite{penza2017envisors} proposed to track the interested soft tissue and then recover the corresponding 3D position information of the soft tissue using stereo reconstruction. The experimental result showed that the AR effect can be implemented on the intra-operative images. Nevertheless, this approach relied on an external device to manually draw the boundary of the interested target at the beginning of the operation, and the pre-operative model was substituted using a simple ellipsoid that is different from the intra-operative tissue.

Many stereo reconstruction methods have been proposed to reconstruct the 3D scene by estimating the disparity map, and the depth information can be recovered by a transformation with the focal length and the baseline of the stereo endoscope. \cite{chang2018pyramid} is a classic stereo reconstruction network, the authors adopted the Convolutional Neural Network (CNN) with embedding a Spatial Pyramid Pooling (SPP) module to extract the high-level features of the stereo image, then they were concatenated as a 4D cost volume and fed into a stacked hourglass architecture for the disparity estimation. To simplify this hourglass architecture, the authors in\cite{guo2019group} adopted fewer skipping connections in the decoder and designed a novel 4D cost volume by calculating group-wise correlation. Next, \cite{yang2019hsm} was proposed to predict image pairs with a high resolution, different-level feature maps were employed to foster the multiple cost volumes and then gradually connected to estimate the disparity map based on a coarse-to-fine manner. More methods~\cite{liu2022graftnet, gu2020cascade, shen2021cfnet} were proposed to adopt different strategies to foster the cost volume since it is a key factor for the final prediction. Different from the traditional CNN architecture, attention-based transformer \cite{vaswani2017attention} provides a new network architecture and vision transformer \cite{dosovitskiy2020image} opens the path to utilize this module in the vision field by encoding the image into many tokens. The works in~\cite{li2021revisiting, cheng2022deep} started to fuse the transformer and CNN for the disparity estimation and their results showed a satisfactory performance by evaluating a public stereo endoscopic dataset~\cite{allan2021stereo}. 

3D reconstruction recovers the depth information of the whole intra-operative scene, which raises another issue, i.e., keeping the interested region while removing other background points for an accurate registration between the pre-operative model and the intra-operative target. Hence, another segmentation network needs to be integrated to distinguish the background and the target by predicting a binary mask. 
UNet~\cite{ronneberger2015u} is the most representative model in medical image segmentation, and it adopted a U-shape architecture with downsampling and upsampling operation and implemented the fusion of different-level features using skipping connection. Following the similar UNet architecture, the authors in~\cite{zhuang2018laddernet} proposed a multi-path feature fusion strategy by combining two UNet networks, the authors in~\cite{oktay2018attention} integrated an attention gate module to enhance the learning of target structures, and the authors in~\cite{zhou2019unet++} designed a nested UNet architecture by densely aggregating different-scale features. The emerging Transformer module has also been utilized in the segmentation field. In~\cite{chen2021transunet, zheng2021rethinking, hatamizadeh2022unetr}, the authors employed the transformers as the encoder to extract the features, and then adopted the CNN architecture as the decoder to predict the segmentation mask. 
More recently, a fancy model named Segment Anything (SAM)~\cite{kirillov2023segment} was released which had been trained based on 11 million images and showed a strong generalization ability in various segmentation tasks. With the explosion of big data, it can be foreseen that such kind of models will lead a new era because of the promising zero-shot performance.

Intra-operative bleeding is a risky situation that affects surgical quality and post-operative recovery of patients, and it generally occurs due to unconscious collisions between the surgical instruments and the delicate blood vessel (i.e., the artery and the vein) during the operation. Hence, we propose a markerless augmented reality framework to relieve the occurrence of this situation. Different from the existing approaches such as the manual location or the pre-installed markers, we combined the stereo reconstruction and segmentation networks to locate the intra-operative soft tissue. The proposed framework integrating the advanced neural networks can implement the visualization of the pre-operative model, and it also performs the minimum distance detection between the instruments and the blood vessel to avoid a dangerous collision. Furthermore, it does not rely on the extra external device, which means high generalization in other robotic systems and surgical tasks. The main contributions can be summarized as follows:

1) A markerless augmented reality framework was proposed to visualize the pre-operative model on intra-operative scenes, and it provides the minimum distance detection between the surgical instruments and the delicate blood vessel for safety. 

2) A comprehensive evaluation of advanced neural networks in stereo reconstruction and segmentation fields was performed to find out the best combination to recover 3D information of the intra-operative blood vessel accurately and fastly. 

3) A user study involving ten human subjects who performed a robot-assisted lymphadenectomy based on the da Vinci Research Kit in a dry lab, was achieved to explore the usability of the proposed AR framework compared to the standard setup.

The remainder of this paper is structured as follows. Section~\ref{sec:Methodology} describes the details of our proposed augmented reality framework. In Section~\ref{Experimental protocol and evaluation metrics}, it presents the framework evaluation metrics and the specific experimental protocol of the usability study, and the results are given in Section~\ref{Results}. Section~\ref{Discussion} discusses the findings and limitations of our work, and the conclusion of this paper and future work are drawn in Section~\ref{CONCLUSIONS}.

\section{Methodology}
\label{sec:Methodology}

\begin{figure*}[!ht] 
    \centering
    \includegraphics[width=1\linewidth]{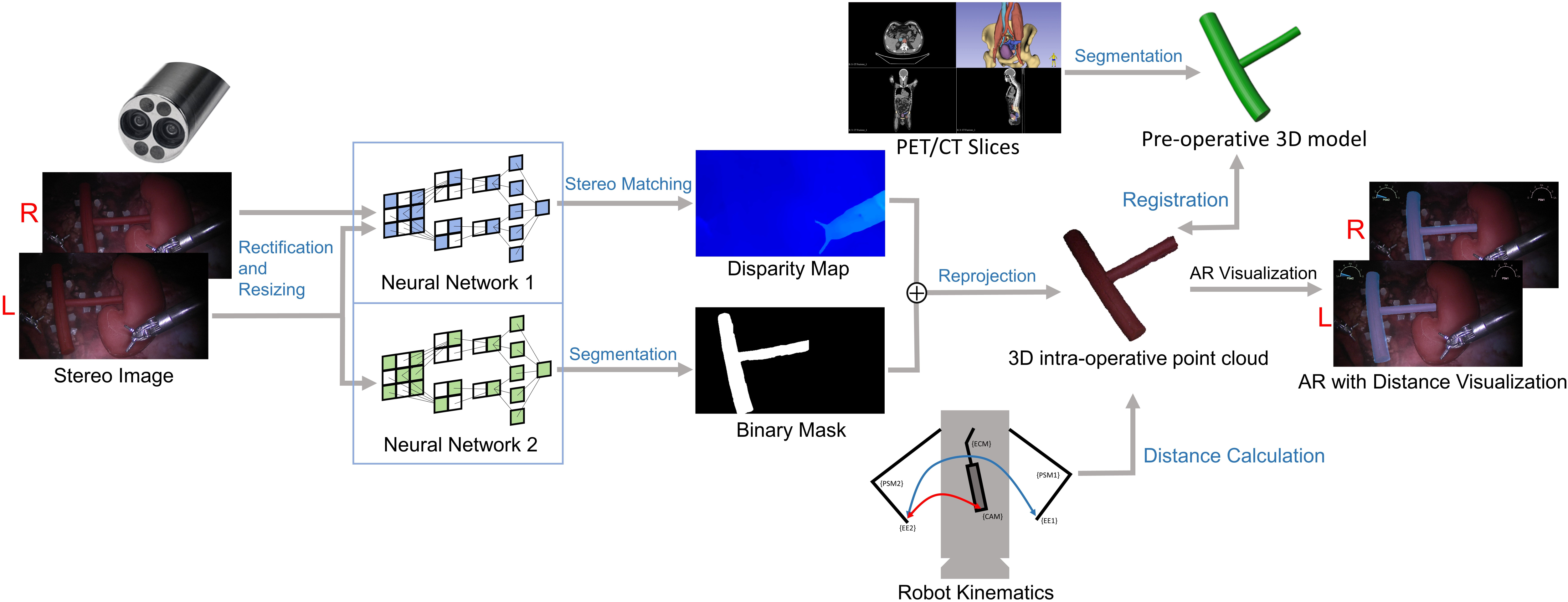}
    \caption{The architecture of the markerless augmented reality framework. The image pair is fed into two different networks to estimate the disparity map and the binary mask. Then, the disparity pixels that belong to the blood vessel can be reprojected to generate the 3D intra-operative blood vessel. The pre-operative model can be overlapped on the corresponding region of the endoscopic images by registration, and the minimum distance between the instruments and the blood vessel can be detected based on the dVRK kinematics and the 3D reconstructed blood vessel. 
}
    \label{fig:pipeline}
\end{figure*}

The proposed augmented reality framework is shown in Fig.~\ref{fig:pipeline}, and it is integrated into the popular da Vinci Research Kit (dVRK, Intuitive Surgical Inc., US, and Johns Hopkins University). The stereo image pair is input into a stereo reconstruction network to estimate the left disparity map, and the left image is also input into a segmentation network to segment the blood vessel. Then, the intra-operative blood vessel can be generated in the 3D space by combining the disparity map and the binary mask. The pre-operative model is utilized to perform the registration with the intra-operative blood vessel so that it can be projected on the corresponding position of the endoscopic image pairs to implement an AR effect. Furthermore, the minimum distances between the surgical instruments and the soft tissue are calculated based on the dVRK kinematics and the 3D position of the reconstructed blood vessel. The specific description of this framework is given below.

\subsection{dVRK system and calibrations}

The dVRK system is an open-source robotic platform composed by the hardware of the first generation of da Vince surgical system as well as the customized software and electronics. It can be mainly divided into the leader side and the follower side. There are two Patient Side Manipulators (PSMs) that mount various surgical instruments such as the large needle driver at the follower side, and a stereo endoscope (1920$\times$1080 resolution) mounted on an Endoscopic Camera Manipulator (ECM) is used to capture in vivo surgical scenes. By adjusting the position of the Setup Joint (SUJ) connecting these robotic arms, the respective Remote Center of Motion (RCM) can be located at the skin entry point of the abdomen~\cite{penza2017envisors}. On the other hand, the surgeon at the leader side can observe the endoscopic scenes using a High Resolution Stereo Viewer (HRSV) and remotely control the movements of PSMs and ECM by operating Master Tool Manipulators (MTMs) as well as a foot pedal tray~\cite{da2020scan, chen2022robot}. 

\begin{figure*}[!ht] 
    \centering
    \includegraphics[width=1\linewidth]{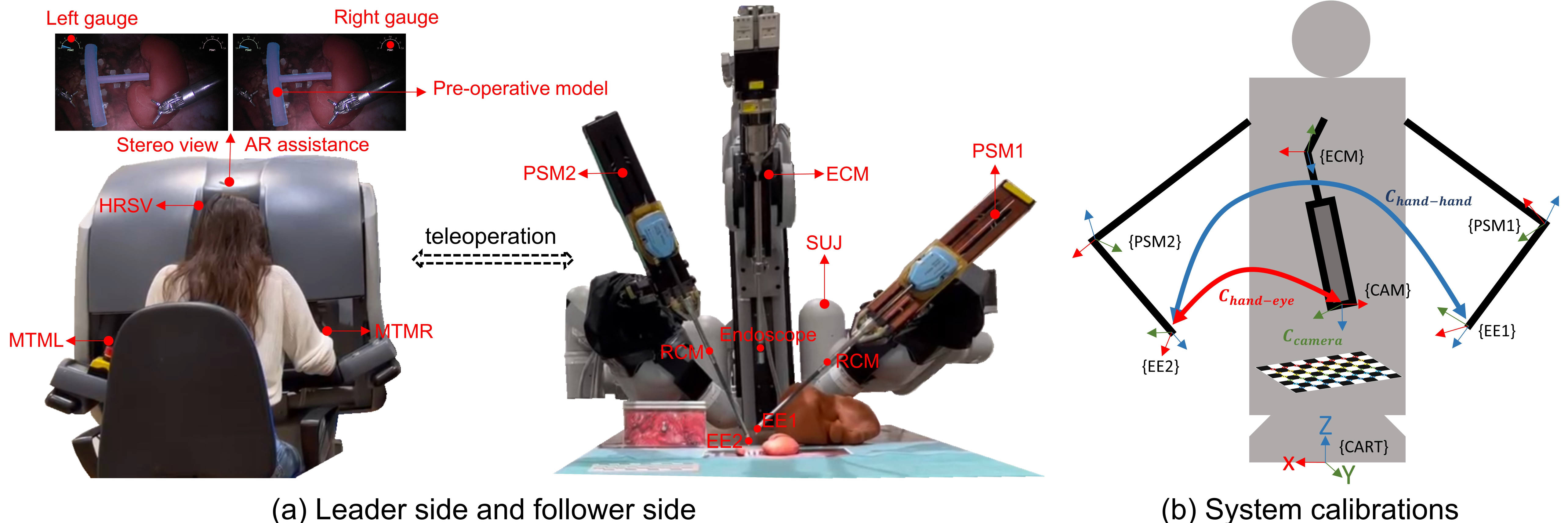}
    \caption{The presentation of the dVRK system in a dry lab. In (a), the user is operating the MTMs and observing the surgical scenes using HRSV at the leader side, and the surgical instruments mounted on the PSMs are performing the operation following the remote control of the user at the follower side. In (b), it shows the three types of system calibrations, including camera calibration, hand-hand calibration and hand-eye calibration.
}
    \label{fig:Kinematics_new}
\end{figure*}

Fig.~\ref{fig:Kinematics_new} shows the details of our dVRK system, and it also presents the reference frame definition adopted in our framework. The Cartesian position of each instrument can be obtained from the dVRK kinematics, and a 9$\times$6 chessboard with a square length of 1cm is adopted for our calibrations. First, the camera calibration is done based on Zhang's calibration approach~\cite{zhang2000flexible} to generate the intrinsic and extrinsic parameters for the image rectification, undistortion and projection.
Then, a hand-to-hand calibration is conducted to search for the rigid transformation between left and right end effectors based on Horn's method~\cite{horn1987closed}, since the position subscribed from the direct kinematics is not accurate enough. By collecting 40 non-collinear points, we can generate the rigid transformation matrix $T_{\emph{EE1}}^{\emph{EE2}}$ and transfer the 3D points in \emph{\{EE1\}} to \emph{\{EE2\}}. Furthermore, a hand-eye calibration is performed to obtain the transformation between the left end effector \emph{\{EE2\}} and the left camera \emph{\{L\_CAM\}}. We operate the left end effector to point at 54 corner points of the chessboard to obtain the 3D coordinates, and the corresponding 2D coordinates on the left image are obtained using cv.findChessboardCorners and cv.cornerSubPix (OpenCV) functions so that the transformation can be calculated based on the Random Sample Consensus (RANSAC) scheme (cv.solvePnPRansac function)~\cite{opencv_library, fischler1981random}.
Since our end effector is referencing the frame \emph{\{ECM\}}, the transformation matrix $T_{\emph{ECM}}^{\emph{L\_CAM}}$ is generated based on the hand-eye calibration.

\subsection{Intra-operative blood vessel reconstruction}

The stereo image is first rectified to align the polar lines in the horizon axis, and the resolution is resized from 1920$\times$1080 to 640$\times$360 to accelerate the framework. Then, the images are fed into a stereo reconstruction network to estimate the disparity map \cite{yang2019hsm}. In this way, we can reproject the disparity map to generate the 3D intra-operative point cloud. The conversion between the estimated disparity value $\hat{d}$ and the depth value $\tilde{d}$ is formulated as,

\begin{equation}
\tilde{d}(i, j)=\frac{b f}{\hat{d}(i, j)}
\end{equation}
where $i$,$j$ are the pixel position on the 2D image, $b$ is the baseline of the stereo camera, and $f$ is the focal length.

However, the reconstructed point cloud involves the whole intra-operative scene, and we need to extract the interested region (i.e., the blood vessel in our case) for the registration. Considering that the estimated disparity map is referencing the left rectified image, we propose to adopt another segmentation network to generate the binary mask of the interested region. The left rectified image is fed into the segmentation network to estimate the indices of the blood vessel and the background. Next, we use the disparity pixels that belong to the blood vessel by referencing these indices and reproject them to the 3D space to reconstruct the intra-operative blood vessel. The 3D position of the reconstructed blood vessel directly influences the quality of the following registration and the distance detection, so a comprehensive comparison study including the stereo reconstruction networks and segmentation networks is provided in Section~\ref{Experimental protocol and evaluation metrics} to determine the best combination in our framework. Finally, two post-processing approaches are adopted to improve the segmentation estimation, including mask boundary eroding and small object removal. Mask boundary eroding can remove the possible misclassified pixels near the boundary of the blood vessel, and the small object removal is used to remove the possible outliers in other regions, which can refine the segmentation quality in some challenging cases such as the blurred scenes caused by the fast movement of the instruments.

The recontructed 3D blood vessel is referencing the rectified left camera coordinate system \emph{\{Rec\_L\_CAM\}} , and we transform it to the reference \emph{\{ECM\}} for the following distance calculation based on the equation,
\begin{equation}
P_{\emph{intra\_obj}}^{\emph{ECM}}=\left(T_{\emph{ECM}}^{\emph{L\_{CAM}}}\right)^{-1}  * \left(T_{\emph{L\_CAM}}^{\emph{Rec\_L\_CAM}}\right)^{-1} * P_{\emph{intra\_obj}}^{\emph{Rec\_L\_CAM}}
\end{equation}
where $P_{\emph{intra\_obj}}^{\emph{Rec\_L\_CAM}}$ is the reconstructed 3D points of the intra-operative blood vessel referencing the rectified left camera coordinate system, $T_{\emph{L\_CAM}}^{\emph{Rec\_L\_CAM}}$ is the transformation between the unrectified left camera system and the rectified left camera system obtained from the stereo image rectification, and $T_{\emph{ECM}}^{\emph{L\_{CAM}}}$ is obtained based on the hand-eye calibration.

\subsection{Registration between pre-operative and intra-operative targets}
The pre-operative 3D model is generally captured using CT or MRI in the hospital, and then using some software such as the 3D Slicer to segment the interested region to generate the 3D structure. In our case, we utilize a 3D modeling software, i.e., Blender, to create a pre-operative 3D model for simplification since we don't have an external device to capture these CT/MRI slices. Next, we can perform the registration between the pre-operative model and the intra-operative reconstructed point cloud. Here, the pre-operative blood vessel is a mesh model while the intra-operative one is a point cloud, so we sample plenty of points from the mesh model for the registration process. Considering that there is an apparent position difference between the pre-operative model and the intra-operative blood vessel in the initial state, the global registration-based RANSAC algorithm~\cite{fischler1981random} is adopted to conduct the initial transformation, and it is only implemented once at the beginning. Then, the local registration-based Iterative Closest Point (ICP)~\cite{besl1992method} algorithm is performed to finetune the position of the pre-operative model. It is found that this registration strategy can register the models accurately and fastly in our experiment.

\subsection{Distance detection and AR visualization}
Our framework not only provides the augmented pre-operative model visualization on intra-operative images, but also detects the minimum distances between the surgical instruments and the blood vessel. After aligning the position of the pre-operative model with the intra-operative blood vessel, we could overlap the pre-operative model on the left intra-operative scenes $P_{\emph{pre\_obj}}^{\emph{L\_img}}$ by the transformation, 

\begin{equation}
P_{\emph{pre\_obj}}^{\emph{L\_img}}=K_\emph{L} * T_{\emph{ECM}}^{\emph{L\_CAM}} * T_{\emph{BL}}^{\emph{ECM}} * P_{\emph{pre\_obj}}^{\emph{BL}}
\end{equation}
where $P_{\emph{pre\_obj}}^{\emph{BL}}$ is the 3D pre-operative points referencing the Blender coordinate system $\emph{\{BL\}}$, $T_{\emph{BL}}^{\emph{ECM}}$ is the transformation matrix generated by the registration, $T_{\emph{ECM}}^{\emph{L\_CAM}}$ is the matrix obtained from the hand-eye calibration and $K_\emph{L}$ contains the intrinsic and distortion matrices of the left camera obtained from the camera calibration. Similarly, we can use the following equation to project the pre-operative model on the right intra-operative scenes $P_{\emph{pre\_obj}}^{\emph{R\_img}}$,

\begin{equation}
P_{\emph{pre\_obj}}^{\emph{R\_img}}=K_\emph{R} * T_{\emph{L\_CAM}}^{\emph{R\_CAM}} * T_{\emph{ECM}}^{\emph{L\_CAM}} * T_{\emph{BL}}^{\emph{ECM}} * P_{\emph{pre\_obj}}^{\emph{BL}}
\end{equation}
where $T_{\emph{L\_CAM}}^{\emph{R\_CAM}}$ is the transformation between the left and right camera coordination systems, and $K_\emph{R}$ contains the intrinsic and distortion matrices of the right camera. Then, we can observe that the pre-operative model is overlapped on the corresponding regions of the left and right images, respectively.

Finally, we calculate the minimum distance between the surface of the instruments and the reconstructed blood vessel based on the fast k-nearest-neighbor search strategy~\cite{point-cloud-utils, garcia2008fast}. The Cartesian position of the end effectors and the RCM points of instruments are subscribed from the robot kinematics so that we can model the instruments as cylinders with a radius of 4mm and sample them as the point clouds (here, the position of PSM1 is aligned to PSM2 by the hand-hand rigid transformation $T_{\emph{EE1}}^{\emph{EE2}}$). Two gauges are provided in the left upper and right upper corners of intra-operative images, and they can visualize the respective minimum distance of left and right instruments. Also, the color of the pre-operative model changes automatically according to the smaller distance by comparing the left and right minimum distances to remind the surgeons during the operation.

\section{Experimental protocol and performance metrics}
\label{Experimental protocol and evaluation metrics}

\subsection{Framework characterisation evaluation}
\subsubsection{Reconstruction and segmentation networks} 

The prerequisite for performing AR visualization and distance detection is that the intra-operative position of soft tissue needs to be accurately restored, so we explored 14 state-of-the-art methods in the stereo reconstruction field to find out the best model that can be utilized in the medical scenes. Among them, ELAS~\cite{geiger2011efficient} is an optimation based method while others~\cite{cheng2022deep, xu2022attention, li2021revisiting, liu2022graftnet, gu2020cascade, chang2018pyramid, guo2019group, shen2021cfnet, garg2020wasserstein, cheng2020hierarchical, xu2023iterative, li2022practical, yang2019hsm} utilize the neural networks. The stereo endoscopic dataset, SERV-CT~\cite{edwards2022serv}, was adopted to conduct the quantitative evaluation of these methods, and it contains sixteen image pairs captured from porcine samples based on the dVSS and provides the dense ground truth. To reconstruct the endoscopic scenes, we run these models based on their official weights without any task-specific finetuning for the generalization~\cite{lu2021super}. A set of accuracy-related metrics was chosen to evaluate the reconstruction error, comparing the estimated depth with the provided ground truth, both expressed in millimeters. The metrics include Median Absolute Error (MeAE), Mean Absolute Error (MAE), Root Mean Square Error (RMSE), Absolute Relative Error (Abs Rel), Squared Relative Error (Sq Rel), as well as $\delta_{ratio}$~\cite{eigen2014depth, zhao2020monocular}.

\begin{equation}
\text{MeAE}= \text{Median} \left\{\tilde{d}(i, j) \in S \mid | \tilde{d}(i, j)-d^{\prime}(i, j) |\right\}
\end{equation}
\begin{equation}
\text{MAE}=\frac{1}{|S|} \sum_{(i, j)}\left|\tilde{d}(i, j)-d^{\prime}(i, j)\right|
\end{equation}
\begin{equation}
\text{RMSE}=\sqrt{\frac{1}{|S|} \sum_{(i, j)}\left|\tilde{d}(i, j)-d^{\prime}(i, j)\right|^{2}}
\end{equation}
\begin{equation}
\text{Abs~Rel}=\frac{1}{|S|} \sum_{(i, j)} \frac{\left|\tilde{d}(i, j)-d^{\prime}(i, j)\right|}{d^{\prime}(i, j)}
\end{equation}
\begin{equation}
\text{Sq~Rel}=\frac{1}{|S|} \sum_{(i, j)} \frac{\left|\tilde{d}(i, j)-d^{\prime}(i, j)\right|^{2}}{d^{\prime}(i, j)}
\end{equation}
\begin{equation}
\delta_{ratio}: \% \text { of } \tilde{d}(i, j) \text { s.t. } \max \left(\frac{\tilde{d}(i, j)}{d^{\prime}(i, j)}, \frac{d^{\prime}(i, j)}{\tilde{d}(i, j)}\right) <\tau
\end{equation}
where $S$ is the set of predicted depth values for each frame, $\tilde{d}(i, j)$ is the predicted depth value related to pixel in position $(i, j)$ and $d^{\prime}(i, j)$ is the ground truth of depth value. The last metric evaluates the depth fluction error between the reconstructed points and the ground truth, and three different thresholds $\tau \in [1.25^{1}, 1.25^{2}, 1.25^{3}]$ were adopted. Unlike the other metrics, the higher $\delta_{ratio}$ means the better reconstruction result. Also, we provided the inference time of one frame when evaluating these models. 

Another segmentation network is required to extract the interested soft tissue, so 8 state-of-the-art segmentation methods~\cite{ronneberger2015u, zhuang2018laddernet, oktay2018attention, zhou2019unet++, chen2021transunet, zheng2021rethinking, hatamizadeh2022unetr, kirillov2023segment} were evaluated for the blood vessel segmentation. The recent segmentation network SAM~\cite{kirillov2023segment} has a strong generalization in different fields, while other neural networks need to be trained. Hence, we captured six endoscopic videos containing the 3D-printed blood vessel based on the dVRK platform in our lab and extracted around 100 images from each video for the manual annotation (551 frames in total). We performed the annotation using Computer Vision Annotation Tool (CVAT)~\cite{sekachev2020opencv}. Six-fold cross validation was adopted to train and evaluate the models. During the training process, we cropped the images into 128$\times$128 patches and trained the models for 100 epochs based on a batch size of 64 (100000 patches in every epoch and 10$\%$ of images from the training images were used for the model validation). Two methods~\cite{chen2021transunet, zheng2021rethinking} loaded the pre-trained weight following the authors' original configuration while other models were trained from the scratch. The images were also split into patches with the resolution of 128$\times$128 without overlapping during the test phase. Common evaluation metrics in the segmentation field were used for a comprehensive comparison, including Dice coefficient ($\frac{2 T P}{2 T P+F P+F N}$), accuracy ($\frac{T P+T N}{T P+T N+F P+F N}$), specificity ($\frac{T N}{T N+F P}$), sensitivity ($\frac{T P}{T P+F N}$), precision($\frac{T P}{T P+F P}$), the area of the precision-recall (PR) curve~\cite{venugopal2022real} as well as the inference time. 

\begin{figure*}[!ht] 
    \centering
    \includegraphics[width=1\linewidth]{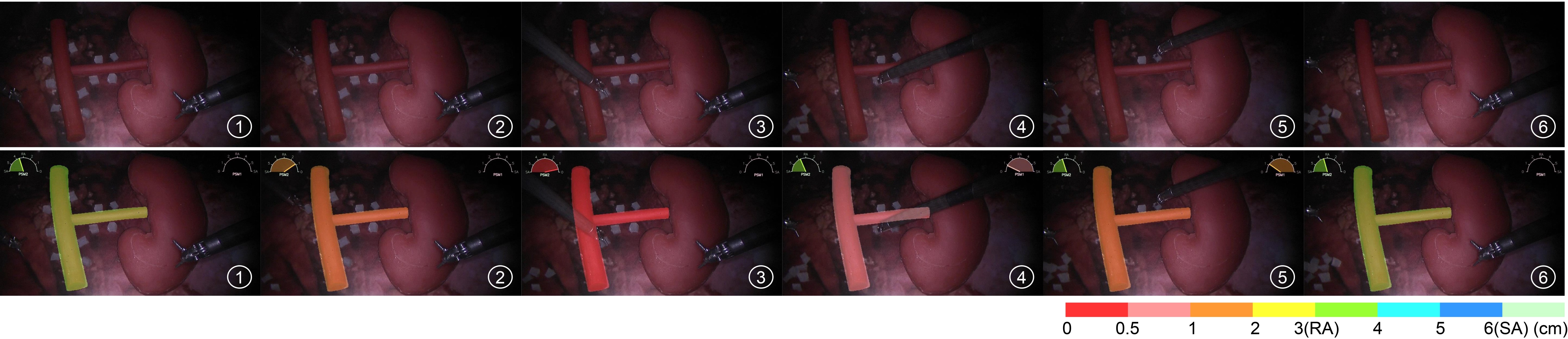}
    \caption{A simulated lymphadenectomy based on the dVRK platform in a dry lab. The first row presents the standard endoscopic scenes, while the second row shows the augmented scenes with AR assistance. The two gauges show the respective minimum distance between the two instruments and the blood vessel. The pre-operative model is overlapping on the intra-operative blood vessel, and its color will change automatically following the smaller distance by comparing the left and right distances. The defined task is to remove the ten lymph nodes while not touching the delicate blood vessel for safety. 
}
    \label{fig:task}
\end{figure*}

\subsubsection{other characterisation}
The quantitative performance of this framework needs to be measured for practicality, and the experimental platform is based on an NVIDIA RTX 3080 GPU in a local laptop. On the one hand, we provided the running time of the proposed framework by calculating the time spent in 100 consecutive frames, and we also gave the specific time distribution in each component of this framework. On the other hand, we calculated the error distribution in different components of this framework. The system calibrations based on the dVRK introduce errors, and the respective performance metric can be formulated as:

$\bullet$ Camera calibration error $E_{\text{cam}}$ considers the difference between the coordinates of reprojected 3D points on the 2D plane and the actual coordinates of the chessboard corners by calculating the root mean square value,

\begin{equation}
E_{\text {cam }}=\sqrt{\frac{1}{N_{\text {cam }}} \sum_{i=1}^{N_{\text {cam }}}\left\|\varepsilon_i^{\text {pro}}-\varepsilon_i^{\text {act}}\right\|_2^2}
\end{equation}
where $\varepsilon_i^{\text {pro}}$ is the 2D points reprojected using the camera calibration parameters, while $\varepsilon_i^{\text {act}}$ denotes the actual 2D coordinates on the chessboard corners. $N_{\text {cam}}$ is the number of the points used for the camera calibration, and $\|,\|_2$ represents the Euclidean norm.

$\bullet$ Hand-hand calibration error $E_{\text {hand\_hand}}$ was calculated based on the Cartesian position difference between the 3D points subscribed from the left end effector and the same points transformed from the right end effector based on the hand-to-hand matrix $T_{\emph{EE1}}^{\emph{EE2}}$,

\begin{equation}
E_{\text {hand\_hand}}=\sqrt{\frac{1}{N_{\text {hh}}} \sum_{i=1}^{N_{\text {hh}}}\left\|\rho_i^L-T_{\emph{EE1}}^{\emph{EE2}} \rho_i^R\right\|_2^2}
\end{equation}
where $\rho_i^L$ and $\rho_i^R$ are the 3D Cartesian positions of the left and right end effectors, respectively. $N_{\text {hh}}$ is the number of the 3D points used for the hand-hand calibration.

$\bullet$ Hand-eye calibration error $E_{\text {hand\_eye}}$ was evaluated based on the pixel position difference between the reprojected 2D coordinates $\gamma_i^{\text {pro}}$ from the 3D points of the left end effector using the hand-eye transformation matrix and the actual 2D coordinates $\gamma_i^{\text {act}}$,

\begin{equation}
E_{\text {hand\_eye}}=\sqrt{\frac{1}{N_{\text {he}}} \sum_{i=1}^{N_{\text {he}}}\left\|\gamma_i^{\text {pro}}-\gamma_i^{\text {act}}\right\|_2^2}
\end{equation}
where $N_{\text {he}}$ is the number of the points adopted for the hand-eye transformation matrix.

Furthermore, the errors of the reconstruction and segmentation networks have been reported in the last subsection, and the registration process also introduces errors. We calculated the Cartesian position error $E_{\text {regis}}$ between the pre-operative blood vessel after registration using the transformation matrix $T_{\emph{BL}}^{\emph{ECM}}$ and the reconstructed blood vessel,

\begin{equation}
E_{\text {regis}}=\sqrt{\frac{1}{N_{\text {re}}} \sum_{i=1}^{N_{\text{re}}}\left\|T_{\emph{BL}}^{\emph{ECM}} \varphi_i^{\text {pre-op }}-\varphi_i^{\text {recon }}\right\|_2^2}
\end{equation}
where $\varphi_i^{\text {pre-op }}$ is the pre-operative 3D points, while $\varphi_i^{\text {recon }}$ is the intra-operative 3D reconstructed points referencing the frame \emph{\{ECM\}}. $N_{\text {re}}$ is the point number for the registration.

\begin{table*}[htbp]
  \centering
  \caption{Quantitative depth estimation result based on the SERV-CT stereo endoscopic dataset (The image resolution is 720$\times$576).}
    \begin{tabular}{cccccccccc}
    \toprule
          & MeAE (mm) & MAE (mm) & RMSE (mm) & Abs Rel & Sq Rel & $\delta<1.25^{1}$ & $\delta<1.25^{2}$ & $\delta<1.25^{3}$ & Time (s) \\
    \midrule
    HybridStereoNet [21] & 46.98±19.32 & 54.30±16.86 & 73.66±13.79 & 0.67±0.08 & 70.27±18.33 & 0.14±0.07 & 0.24±0.12 & 0.39±0.21 & 0.52±0.01 \\
    ACVNet [40] & 8.43±13.46 & 17.88±12.24 & 28.26±14.52 & 0.20±0.12 & 10.12±8.65 & 0.69±0.19 & 0.72±0.19 & 0.77±0.20 & 0.42±0.01 \\
    STTR [20] & 3.82±6.81 & 15.65±10.24 & 42.26±22.28 & 0.22±0.20 & 31.88±33.58 & 0.83±0.18 & 0.88±0.14 & 0.94±0.06 & 0.43±0.00 \\
    ELAS [39] & 3.81±3.14 & 9.05±8.51 & 18.86±19.72 & 0.13±0.16 & 11.73±29.87 & 0.87±0.18 & 0.95±0.07 & 0.98±0.04 & 0.05±0.01 \\
    GraftNet [15] & 1.27±0.55 & 8.23±4.59 & 35.42±19.45 & 0.11±0.08 & 23.43±21.61 & 0.96±0.03 & 0.96±0.03 & 0.97±0.02 & 0.33±0.00 \\
    Cascade-Stereo [16] & 1.80±0.67 & 5.53±3.00 & 13.15±6.72 & 0.06±0.02 & 2.15±1.72 & 0.93±0.04 & 0.96±0.03 & 0.98±0.02 & 0.34±0.00 \\
    PSMNet [12] & 1.43±0.61 & 3.43±1.94 & 7.13±4.81 & 0.04±0.02 & 0.72±1.05 & 0.97±0.03 & 0.98±0.02 & 0.99±0.02 & 0.40±0.01 \\
    GwcNet [13] & 1.77±0.56 & 3.06±1.25 & 5.22±2.77 & 0.04±0.01 & 0.35±0.39 & 0.98±0.03 & 1.00±0.01 & 1.00±0.00 & 0.31±0.01 \\
    CFNet [17] & 1.29±0.49 & 3.04±2.13 & 6.99±7.21 & 0.04±0.05 & 2.01±6.16 & 0.98±0.02 & 0.99±0.02 & 1.00±0.01 & 0.29±0.00 \\
    W-Stereo-Disp [41] & 1.35±0.41 & 3.02±1.38 & 6.59±4.35 & 0.04±0.02 & 0.71±1.04 & 0.98±0.03 & 0.99±0.02 & 1.00±0.01 & 0.42±0.00 \\
    LEAStereo [42] & 1.44±0.82 & 2.96±1.28 & 6.29±2.60 & 0.04±0.01 & 0.58±0.44 & 0.98±0.02 & 1.00±0.01 & 1.00±0.00 & 0.52±0.01 \\
    IGEV-Stereo [43] & 1.12±0.47 & 2.59±1.18 & 5.64±3.10 & 0.03±0.01 & 0.40±0.36 & 0.98±0.02 & 1.00±0.01 & 1.00±0.00 & 0.32±0.00 \\
    CREStereo [44] & 1.32±0.71 & 2.38±1.58 & 4.18±2.90 & 0.03±0.01 & 0.23±0.27 & 0.99±0.01 & 1.00±0.00 & 1.00±0.00 & 1.14±0.01 \\
    HSM [14] & 1.10±0.60 & 2.05±1.17 & 3.72±2.08 & 0.02±0.01 & 0.17±0.17 & 0.99±0.02 & 1.00±0.00 & 1.00±0.00 & 0.03±0.00 \\
    \bottomrule
    \end{tabular}%
  \label{tab:reconstruction}%
\end{table*}%

\begin{figure*}[!ht] 
    \centering
    \includegraphics[width=1\linewidth]{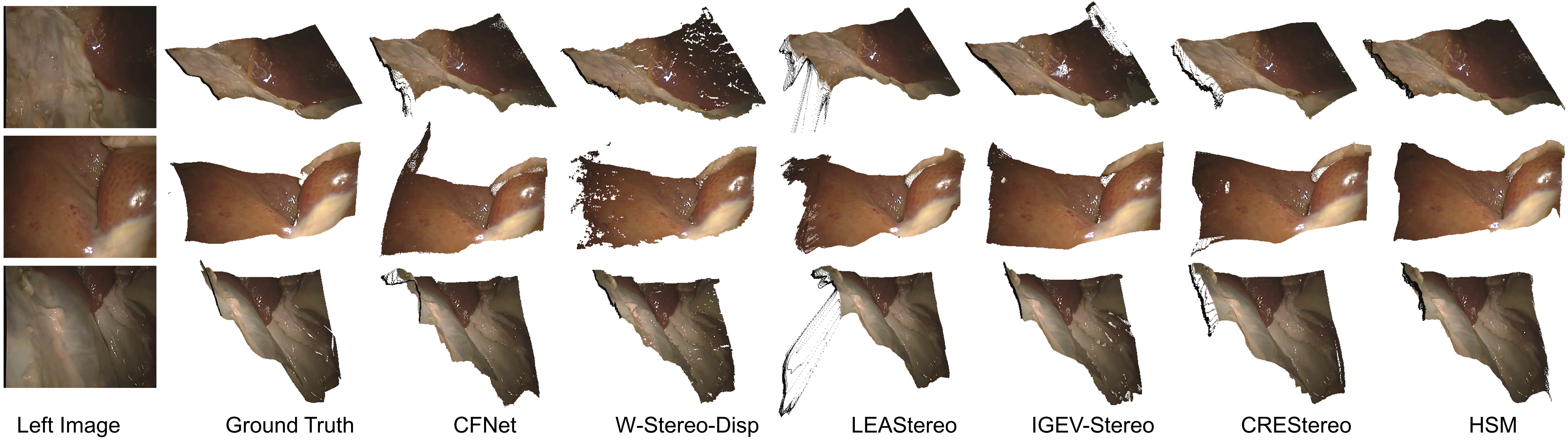}
    \caption{Qualitative surgical scene reconstruction result. The reconstructed 3D surgical surfaces based on six representative models are provided.}
    \label{fig:Serv_CP_comparison}
\end{figure*}

\subsection{Framework usability study}

To explore the usability of our proposed framework, a common surgical operation named lymphadenectomy was designed in a dry lab environment based on the proposal of an oncology surgeon. As shown in Fig.~\ref{fig:task}, the 3D-printed soft blood vessel and kidney were adopted to simulate the surgical scene, and the defined task is to remove the ten lymph nodes one by one (the white soft objects) near the blood vessel while not touching it. Frame 1 shows the initial position of the instruments, then the users operate the left manipulator to catch the left six lymph nodes and put them at the bottom-left corner as shown in Frames 2 and 3, next the users operate the right manipulator to remove the right four lymph nodes and put them at the same corner as shown in Frame 4 and 5, finally the users move the instrument to the initial position in Frame 6. Frame 3 and Frame 5 are challenging cases since the instruments are more likely to collide with the blood vessel. Ten human subjects (6 males and 4 females, aged between 21 and 28) who have a biomedical background were invited to join our experiment. Before the experiment, the users had spent 10 minutes for them to be familiar with the dVRK system. Then, the experiment was repeated in three rounds and we analyzed the user data of the last round to avoid the possible influence of the learning curve. In each round, two different modalities were performed in a random sequence for each user: Control (the standard endoscopic scene without the AR assistance), and Experiment (the endoscopic scene with the AR assistance). 

Our AR framework not only visualizes the corresponding pre-operative model on the intra-operative blood vessel, but it also provides the visualization of the minimum distances between the instruments and the blood vessel. We added two gauges on the scenes to visualize the minimum distances (SA: 6cm means the safe area and RA: 3cm means the risk area), and the color of the pre-operative model changes following the smaller distance between the left and right distances. Five performance metrics were utilized to observe if the AR framework can help improve the surgical performance,

$\bullet$ Minimum distance $D_{\text {min}}$ between the instruments and the blood vessel:
    \begin{equation}
D_{\text {min}}=\min \left\{d_{1L}, d_{2L}, \cdots, d_{ML}, d_{1R}, d_{2R}, \cdots, d_{MR}\right\}
\end{equation}
where $d_{ML}$ is the minimum distance of the left instrument in the M-th frame, while $d_{MR}$ denotes the minimum distance of the right instrument in the M-th frame during the operation.

$\bullet$ Mean distance $D_{\text {mean}}$ when the instruments are in the risk area of 3cm:
\begin{equation}
D_{\text {mean}}=\frac{1}{\widetilde{M}+\widetilde{N}}\left(\sum_{m=1}^{\widetilde{M}} d_{m L}+\sum_{n=1}^{\widetilde{N}} d_{n R}\right)
\end{equation}
where $\widetilde{M}$ is the number of the minimum distance points of the left instrument when they are smaller than 3cm, while $\widetilde{N}$ represents the number when the points of the right instrument are less than 3cm.

$\bullet$ Collision number $N_{\text {c}}$ when the distance points are smaller than the threshold $r$:
\begin{equation}
N_{\text {c}}=\sum_{m=1}^M\left\{1 \mid d_{m L}<r\right\}+\sum_{m=1}^M\left\{1 \mid d_{mR}<r\right\}
\end{equation}
where $r$ is defined as 0.5cm in our case. Here, we regard it as one time of collision if the points keep smaller than 0.5cm for one consecutive second during the task.

$\bullet$ Overall movement path $S_{\text {p}}$ of the instruments during the operation:

\begin{equation}
S_p=\sum_{m=2}^M\left(\left\|C_L^m-C_L^{m-1}\right\|_2+\left\|C_R^m-C_R^{m-1}\right\|_2\right)
\end{equation}
where $C_L^m$ is the 3D Cartesian coordinate of the left end effector in the m-th frame, while  $C_R^m$ is the 3D coordinate of the right one in the m-th frame.

$\bullet$ Execution time $T_{\text {exe }}$ to perform the complete operation:
\begin{equation}
T_{\text {exe }}=T_{\text {end }}-T_{\text {start }}
\end{equation}
where $T_{\text {start }}$ and $T_{\text {end }}$ are the start time and the end time of the task, respectively.

\begin{table*}[htbp]
  \centering
  \caption{Quantitative segmentation result using the self-made dataset (resolution: 1920$\times1080$) based on 6-fold cross validation.}
    \begin{tabular}{cccccccc}
    \toprule
          & DSC   & Accuracy & Specificity & Sensitivity & Precision & Area of PR Curve & Time (s) \\
    \midrule
    Attention UNet [25] & 0.6708±0.2296 & 0.9089±0.0778 & 0.9059±0.0856 & 0.9486±0.0404 & 0.5860±0.3075 & 0.7974±0.1324 & 0.3509±0.0018 \\
    Segment Anything [30] & 0.9661±0.0372 & 0.9957±0.0041 & 0.9997±0.0003 & 0.9404±0.0633 & 0.9958±0.0049 & N/A   & 0.6219±0.0071 \\
    UNETR [29] & 0.9780±0.0082 & 0.9971±0.0010 & 0.9986±0.0006 & 0.9757±0.0116 & 0.9805±0.0093 & 0.9931±0.0063 & 0.2056±0.0014 \\
    SETR [28] & 0.9794±0.0076 & 0.9973±0.0009 & 0.9989±0.0004 & 0.9747±0.0126 & 0.9843±0.0064 & 0.9933±0.0068 & 0.1970±0.0010 \\
    TransUNet [27] & 0.9823±0.0059 & 0.9976±0.0006 & 0.9988±0.0006 & 0.9818±0.0101 & 0.9829±0.0085 & 0.9985±0.0015 & 0.3066±0.0029 \\
    UNet++ [26] & 0.9825±0.0062 & 0.9976±0.0007 & 0.9986±0.0007 & 0.9842±0.0080 & 0.9809±0.0099 & 0.9984±0.0017 & 0.3498±0.0016 \\
    LadderNet [24] & 0.9835±0.0061 & 0.9978±0.0007 & 0.9989±0.0005 & 0.9824±0.0100 & 0.9846±0.0071 & 0.9987±0.0017 & 0.0898±0.0012 \\
    UNet [23] & 0.9855±0.0038 & 0.9980±0.0005 & 0.9990±0.0004 & 0.9842±0.0072 & 0.9869±0.0059 & 0.9990±0.0010 & 0.3178±0.0017 \\
    \bottomrule
    \end{tabular}%
  \label{tab:segmentation}%
\end{table*}%

\begin{figure*}[!ht] 
    \centering
    \includegraphics[width=1\linewidth]{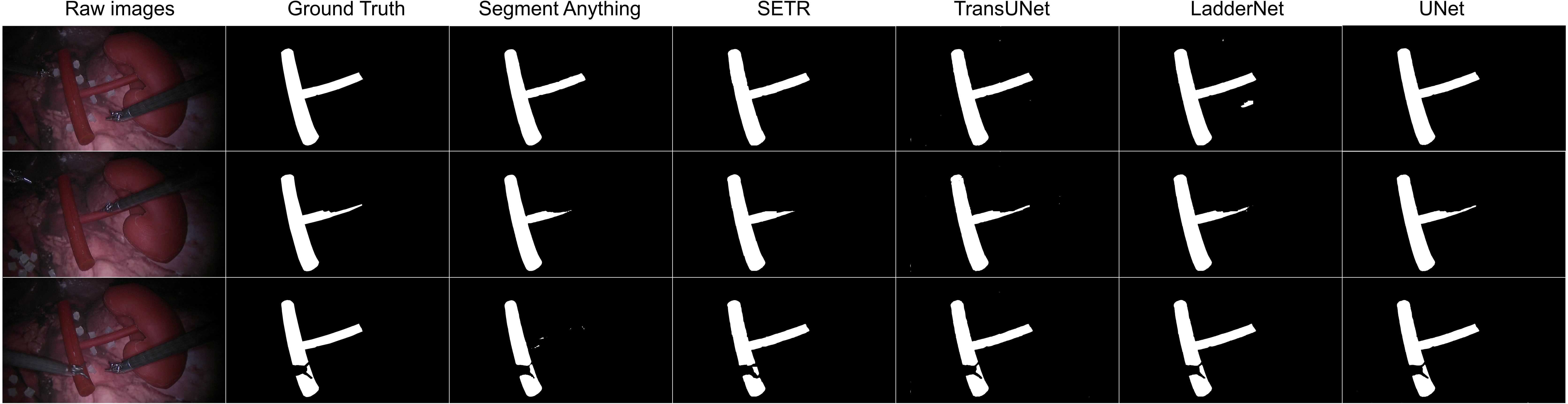}
    \caption{Qualitative segmentation result using a self-made dataset containing the 3D-printed blood vessel captured from the dVRK platform. 
}
    \label{fig:seg_ourlab}
\end{figure*}

To investigate the feasibility and friendliness of the AR framework, the users were invited to fill in a System Usability Scale (SUS) questionnaire containing ten typical questions~\cite{brooke1996sus} after the experiment. They evaluated the two systems (the standard system and the extended system with the AR assistance) by giving a score for each question (from score 1: strongly disagreement to score 5: strongly agreement), and the final SUS score of each user can be calculated as,

\begin{equation}
\text{SUS}_{\text {score }}=\left(\sum_{k=1,3,5,7,9}\left(S_k-1\right)+\sum_{k=2,4,6,8,10}\left(5-S_k\right)\right) * 2.5
\end{equation}
where $S_{k}$ denotes the score of the k-th question provided by the user. 

\begin{table}[t!]
  \centering
  \caption{The running time distribution of the framework (the image resolution is resized into 640$\times$360)}
    \begin{tabular}{cc}
    \toprule
    Phase & Time (s) \\
    \midrule
    Stereo image preprocessing & 0.0095±0.0017 \\
    Disparity map estimation & 0.0365±0.0018 \\
    Binary mask estimation with postprocessing  & 0.0470±0.0015 \\
    Point cloud generation and alignment & 0.0032±0.0004 \\
    Distance calculation & 0.0158±0.0020 \\
    Registration between pre-op and intra-op targets & 0.0015±0.0002 \\
    Augmented reality visualization & 0.0312±0.0043 \\
    \midrule
    Whole pipeline & 0.1448±0.0079 \\
    \bottomrule
    \end{tabular}%
  \label{tab:framework_time}%
\end{table}%

\begin{table}[t!]
  \centering
  \caption{The error distribution in the framework.}
    \begin{tabular}{ccc}
    \toprule
    Component & Metric & Value / Unit \\
    \midrule
    Camera calibration & $E_{\text{cam}}$  & 0.60±0.17 (pixels) \\
    Hand-hand calibration & $E_{\text {hand\_hand}}$ & 0.10±0.05 (cm) \\
    Hand-eye calibration & $E_{\text {hand\_eye}}$ & 1.64±0.80 (pixels) \\
    Reconstruction & MAE & 0.205±0.117 (cm) \\
    Segmentation & Dice  & 0.9855±0.0038 \\
    Registration & $E_{\text {regis}}$ & 0.4328±0.0385 (cm) \\
    \bottomrule
    \end{tabular}%
  \label{tab:error_system}%
\end{table}%

\section{Results}
\label{Results}
\subsection{Results on framework characterisation evaluation}
\subsubsection{Reconstruction and segmentation networks}
Table~\ref{tab:reconstruction} gives the quantitative evaluation result using these advanced depth estimation models. The model HSM\cite{yang2019hsm} gets the best performance in both accuracy and inference time. The qualitative comparison based on three frames was also provided in Fig.~\ref{fig:Serv_CP_comparison}. It can be observed that HSM can reconstruct smoother soft tissue surfaces with fewer outliers, so we chose this method for the depth estimation in our framework. 

The quantitative segmentation result was provided in Table~\ref{tab:segmentation}, and the qualitative comparison result was shown in Fig.~\ref{fig:seg_ourlab}.
Here, we provided the specific boxes as the prompt for the mask estimation of the SAM model, and the area of PR curve is not applicable for this model since it predicts the pixel classification directly instead of the probability. We noticed that UNet could provide a reliable segmentation quality in our phantom environment, so we adopted this model to estimate the mask in our experiment. 

\begin{figure*}[!ht] 
    \centering
    \includegraphics[width=1\linewidth]{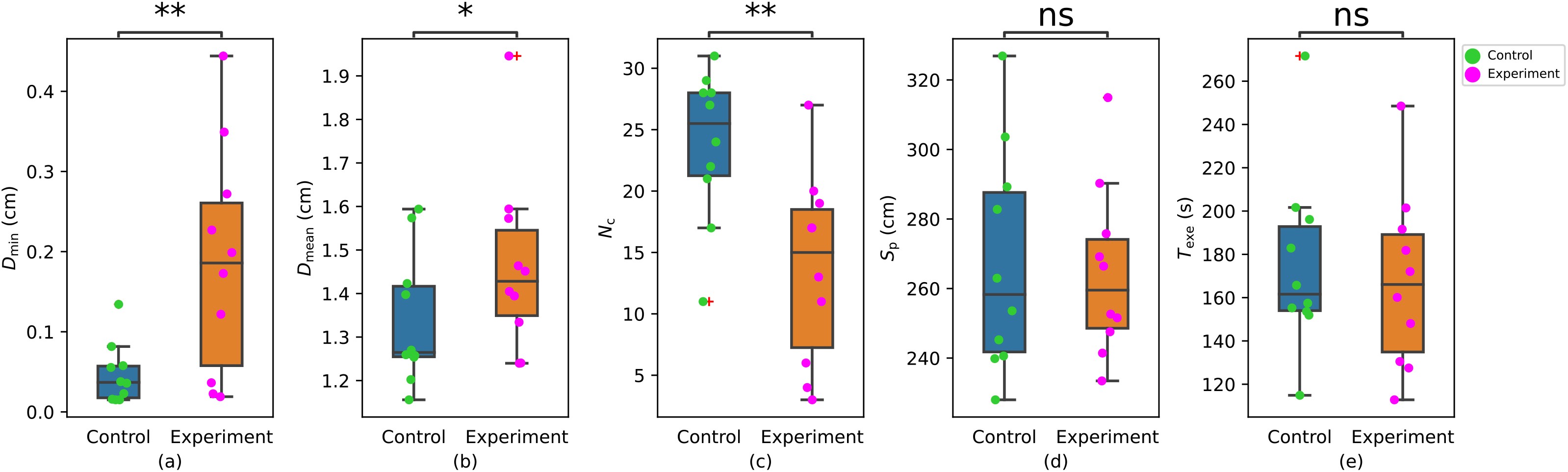}
    \caption{The data of the ten human subjects based on five different metrics. ``Control'' means the users complete the operation based on the standard endoscopic scenes, while ``Experiment'' is based on the scenes with the AR assistance.
    The result of the Wilcoxon signed-rank test is shown as $ns: 0.05 < p \leq 1$, $*: 0.01 < p \leq 0.05$, $**: 0.001 < p \leq 0.01$, $***: 0.0001 < p \leq 0.001$, and $***~*: p \leq 0.0001$. 
}
    \label{fig:user_result}
\end{figure*}

\subsubsection{Other characterisation}

Table~\ref{tab:framework_time} presents the running time distribution of this framework. Among them, stereo image preprocessing includes image subscribing, rectification and resizing from 1920$\times$1080 to 640$\times$360 based on a fixed scale factor. It can be noticed that the three phases, i.e., the disparity map estimation, the binary mask estimation and the AR visualization, spent the most time in this framework, and the whole time to process one frame takes 0.1448±0.0079 (6.91FPS), which can provide a smooth visual feedback during the operation. Also, Table~\ref{tab:error_system} shows the error distribution in this framework. It can be noted that these errors are small enough in our experiment.

\subsection{Results on framework usability study}
Fig.~\ref{fig:user_result} presents the box plots of the user data, and the Wilcoxon signed-rank test (p$<$0.05) is conducted to explore if there is a significant difference between the control modality and the experiment modality. It shows that there are significant differences when referencing the minimum distance $D_{\text {min}}$, the mean distance $D_{\text {mean}}$ and the collision number $N_{\text {c}}$. Table~\ref{tab:user_data} also provides the average values using the five metrics. With the AR assistance, the minimum distance increases from 0.0472cm to 0.1864cm, and the mean distance in the risk area also increases from 1.3387cm to 1.4641cm. When considering the collision number, the value reduces from 23.8 to 13.7. The statistical differences show that the AR assistance can help reduce the collision risk between the instruments and the blood vessel when operating. Furthermore, there is no statistical difference when referencing the overall path $S_{\text {p}}$ and execution time $T_{\text {exe }}$, which indicates that the AR assistance does not introduce an extra load in operating the robot.

\begin{figure}[!t] 
    \centering
    \includegraphics[width=1\linewidth]{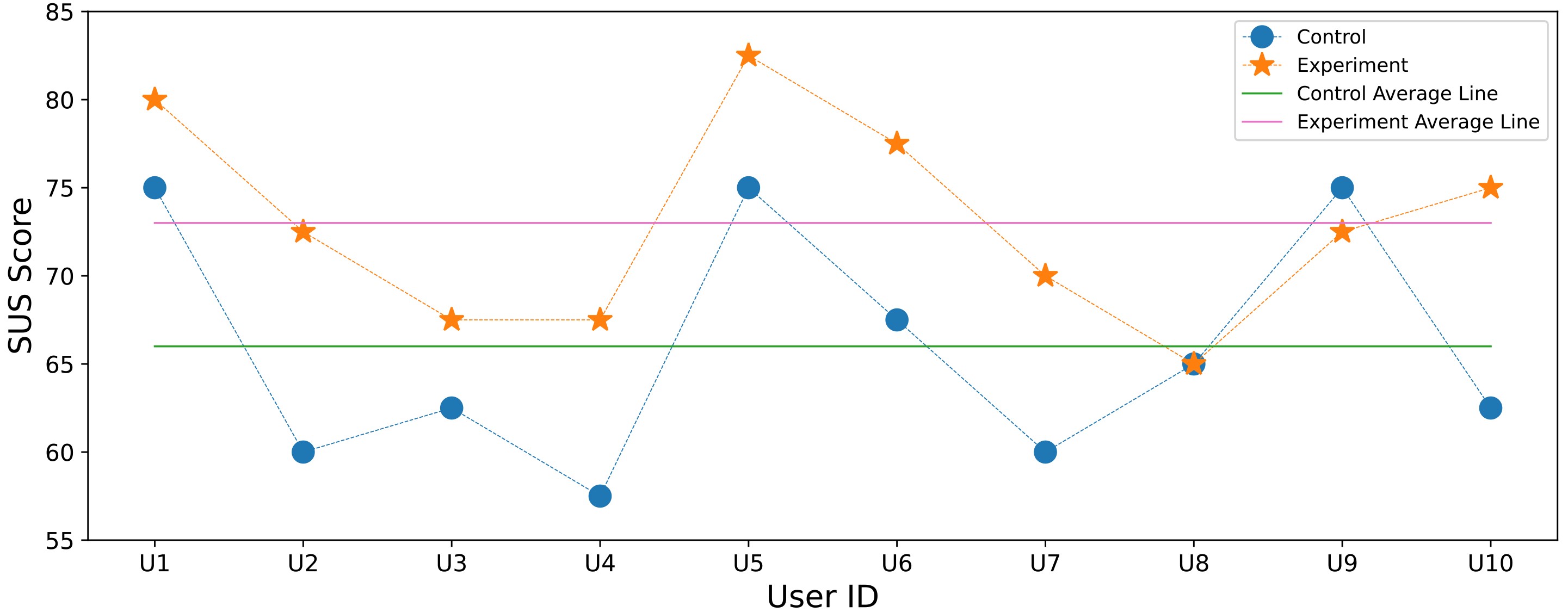}
    \caption{The specific SUS score distribution provided by the users as well as the average SUS scores in two different modalities.
}
    \label{fig:sus_result}
\end{figure}

The final result of the SUS questionnaire is given in Fig.~\ref{fig:sus_result}. The average SUS score of the control modality is 66, while the experiment modality has a higher SUS score of 73. The statistical test presents there is a significant difference between the two modalities (p = 0.0104).

\begin{table}[!t]
  \centering
  \caption{The average values and p value based on the users' data.}
    \begin{tabular}{cccc}
    \toprule
          & Control & Experiment & P value \\
    \midrule
    Minimum Distance $D_{\text {min}}$ (cm) & 0.0472 & 0.1864 & 0.0098 \\
    Mean Distance $D_{\text {mean}}$ (cm) & 1.3387 & 1.4641 & 0.0371 \\
    Collision Number $N_{\text {c}}$ & 23.8  & 13.7  & 0.0077 \\
    Overall Path $S_{\text {p}}$ (cm) & 267.28 & 264.32 & 0.5566 \\
    Execution Time  $T_{\text {exe }}$ (s) & 175.11 & 167.44 & 0.2324 \\
    \bottomrule
    \end{tabular}%
  \label{tab:user_data}%
\end{table}%

\section{Discussion}
\label{Discussion}
Augmented reality is a popular direction in various fields, including robotic surgery since it provides the possibility to enhance safety during the operation, such as the distance visualization in our case or the visualization of some invisible tissues during the operation. However, an unsolved challenge is to locate the region of interest so that the pre-operative model can be registered on the corresponding intra-operative soft tissues or organs. In this work, we proposed a vision-based markerless location approach to perform the AR effect with the distance visualization on the intra-operative scenes, which releases the burden of the manual alignment or landmarks. It can be integrated into other tasks and platforms because of the high independence on the specific device.

Advanced neural networks have been proposed for different vision tasks with promising performance by evaluating some public or self-made datasets, and it is worth integrating them to implement applications in practice. Hence, we compared the state-of-the-art networks in the stereo reconstruction and segmentation fields to track the soft tissue in the 3D intra-operative space. We adopted the model HSM \cite{yang2019hsm} in our case since it shows a reliable depth estimation in both depth and inference time. Furthermore, we added a segmentation mask to extract the region of interest from the whole scene. The segmentation performance relies on the specific scenes, so we captured and annotated an endoscopic dataset based on the dVRK platform in our lab. By referencing the segmentation results of 6-fold cross validation, we noticed that UNet~\cite{ronneberger2015u} can provide a more reliable segmentation quality compared to other models, so we adopted it in our framework. But it should be noted that the segmentation quality is influenced by different scenes and training strategies. For instance, the transformer-based networks~\cite{chen2021transunet, zheng2021rethinking, hatamizadeh2022unetr} may produce a better estimation with a large of annotated training images, which needs to be compared in the specific tasks. In particular, we evaluated the recently emerging big model SAM~\cite{kirillov2023segment}. Although it does not perform a better result compared to some other models, it should be pointed out that this model has a strong generalization ability in different scenes without finetuning. With the rise of big models and data, it can be foreseen that such kind of model may dominate the vision tasks since it releases the burden of annotation and training.

By simulating a robot-assisted lymphadenectomy based on the dVRK platform, we obtained some surgical data from ten human subjects. It can be observed that the proposed AR framework can increase the distance and reduce the collisions between the instruments and the delicate blood vessel when operating. Moreover, it does not introduce the extra physical and cognitive load since the overall path and the complete time are similar in two different modalities. Based on the feedback from the users, sometimes they could not catch the lymph nodes well caused by the inaccurate depth perception when they operated. Under this circumstance, the color information of the pre-operative model can assist them in judging the proper occasion to catch the lymph nodes, especially the pink and red colors (the distances are within 1cm and 0.5cm, respectively), which can enhance their confidence and accuracy when catching the objects. Finally, the average SUS score given by the users increases by 7\% when adopting the AR assistance, which means the proposed AR framework is friendly to the users. Nevertheless, one limitation comes from the modeling of surgical instruments. There are many types of instruments in real surgery, such as monopolar scissors, bipolar forceps, Cadiere forceps, clip appliers and needle drivers, which means the grippers of the instruments are different, and modeling them into a cylinder introduces slight errors. A possible solution is to add an extra neural network to recognize the tips of the instruments \cite{moccia2020vision} so that the modeling can be more accurate even though it may slow down the framework. Another limitation comes from the phantom environment for the simulated surgical operation. In clinical practice, the in vivo environment of patients will be more complex and dynamic.


\section{CONCLUSION}
\label{CONCLUSIONS}
This paper proposed a markerless augmented reality framework to implement a pre-operative structure visualization on the intra-operative scenes, and it also provides the minimum distance detection between the instruments and the delicate blood vessel for safety. It can be integrated into other existing robotic platforms and tasks because it does not rely on specific devices. Comprehensive comparison studies are performed to explore the best combination for intra-operative blood vessel reconstruction, and the framework usability evaluation from ten human subjects presents that the proposed framework can enhance safety during the operation and does not introduce extra burden, which shows its potential in clinical applications. 

Augmented reality provides the possibility to enhance surgical safety based on visual feedback, and another popular direction, virtual fixtures, can also enhance safety by providing force feedback. In the next step, we will conduct the virtual fixtures based on the dVRK platform and compare the difference between the visual feedback and the force feedback in surgical assistance.





\balance
\bibliographystyle{IEEEtran}
\bibliography{ref}

\end{document}